\documentclass[10pt,twocolumn,letterpaper]{article}
\usepackage[letterpaper,textwidth=6.875in,textheight=9in,top=0.8in,columnsep=0.3125in]{geometry}
\usepackage{times}
\usepackage{microtype}
\usepackage{xspace}
\usepackage[dvipsnames]{xcolor}
\usepackage{graphicx}
\usepackage{amsmath,amssymb}
\usepackage{booktabs,multirow}
\usepackage[numbers,sort&compress]{natbib}
\usepackage[format=plain,labelformat=simple,labelsep=period,font=small]{caption}
\usepackage[font=footnotesize,skip=3pt,subrefformat=parens]{subcaption}
\usepackage[shortlabels,inline]{enumitem}
\definecolor{linkblue}{rgb}{0.21,0.49,0.74}
\usepackage[breaklinks,colorlinks,allcolors=linkblue]{hyperref}
\usepackage[capitalize]{cleveref}
\hypersetup{pdftitle={Reconstruction-Guided Dual-Branch Learning for LiDAR Anomalous Object Detection},pdfauthor={Guanyu Jiang, Junjie Cao, Jinshan Pan, Zhixun Su},pdfsubject={Preprint}}

\newcommand{\unit}{\operatorname{n}}

\newcommand{\E}{\mathbb{E}}
\title{\Large\bfseries Reconstruction-Guided Dual-Branch Learning\\for LiDAR Anomalous Object Detection}
\author{\normalsize\textit{Guanyu Jiang\textsuperscript{1}, Junjie Cao\textsuperscript{1}\thanks{Corresponding author.}, Jinshan Pan\textsuperscript{2}, Zhixun Su\textsuperscript{1}}\\[5pt]
\small\textsuperscript{1}Dalian University of Technology (dlut.edu.cn)\\[2pt]
\small\textsuperscript{2}Nanjing University of Science and Technology (njust.edu.cn)}
\date{}
\makeatletter
\newcommand{\experimentparagraph}{\@startsection{paragraph}{4}{\z@}{4pt plus 1pt minus 1pt}{-0.6em}{\normalfont\normalsize\bfseries}}

\renewcommand{\section}{\@startsection{section}{1}{\z@}{-10pt plus -2pt minus -2pt}{7pt}{\normalfont\large\bfseries}}
\renewcommand{\subsection}{\@startsection{subsection}{2}{\z@}{-8pt plus -2pt minus -2pt}{5pt}{\normalfont\fontsize{11}{13}\selectfont\bfseries}}
\renewcommand{\subsubsection}{\@startsection{subsubsection}{3}{\z@}{-6pt plus -2pt minus -2pt}{3pt}{\normalfont\normalsize\bfseries}}
\renewcommand{\@maketitle}{\newpage\null\vskip 8pt
  \begin{center}
    {\@title\par}\vskip 12pt
    {\lineskip .5em\begin{tabular}[t]{c}\@author\end{tabular}\par}
    \vskip 8pt
  \end{center}}
\renewenvironment{abstract}{\begin{center}\large\bfseries Abstract\end{center}\noindent\ignorespaces}{\par\vspace{8pt}}
\makeatother
\begin{document}
\raggedbottom
\hfuzz=0pt\vfuzz=0pt
\maketitle
\suppressfloats[t]

\begin{figure}[t]
\centering
\includegraphics[width=\columnwidth]{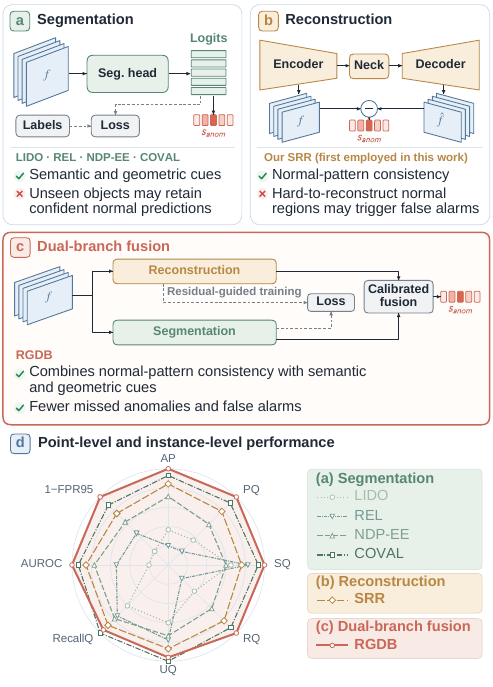}
\caption{\textbf{Paradigms and performance for LiDAR anomalous object detection.} (a) Segmentation uses semantic and geometric cues. (b) SRR assesses normal-pattern consistency. (c) RGDB couples these complementary criteria through reconstruction-guided training and calibrated fusion. (d) Point- and instance-level performance: RGDB outperforms COVAL on most metrics. Radar radii show transformed relative performance; exact values appear in \cref{tab:comparison}.}
\label{fig:teaser_right_column}
\end{figure}

\begin{abstract}
Detecting unexpected objects is essential for autonomous driving safety. Most existing LiDAR anomaly detectors rely on semantic discrimination, which primarily captures category-discriminative features but may overlook deviations from broader feature relationships in normal scenes. Reconstruction offers a complementary perspective by assessing consistency with learned normal patterns. However, pointwise reconstruction errors alone struggle to distinguish anomalies from normal variability in complex outdoor point clouds. We propose Structured Residual Reasoning (SRR) to extract additional anomaly evidence from residual structure across feature channels, time sequences, and geometric space. We further establish that semantic discrimination and normal-pattern reconstruction capture fundamentally different yet complementary anomaly evidence: category compatibility and consistency with learned normal patterns, respectively. At matched background false-positive rates, SRR recovers more objects missed entirely by the semantic branch while producing fewer false positives in normal-only scenes. Building on this finding, we propose Reconstruction-Guided Dual-Branch (RGDB) learning, in which SRR-derived normal-pattern consistency cues guide a semantic-conditioned anomaly decoder to complement reconstruction-based predictions during training. At inference, the two branches jointly exploit their complementary anomaly evidence. Experiments show that standalone SRR establishes a strong reconstruction-based baseline without semantic labels from the target-scene training data, while RGDB achieves state-of-the-art performance on the STU validation set.

\end{abstract}
\section{Introduction}
Unexpected objects on roads threaten autonomous driving safety. LiDAR provides their 3D geometry, but sparse sampling, occlusion, and diverse obstacle shapes make them difficult to detect. LiDAR anomalous object detection (LiDAR AOD) aims to identify points belonging to unexpected objects and recover their instance extents in driving scenes.

Recent methods for LiDAR AOD build on segmentation models, including LIDO~\cite{lido}, REL~\cite{rel}, NDP~\cite{ndp}, and COVAL~\cite{coval}. They identify anomalies from semantic responses, learned anomaly logits, or relationships among category predictions. However, the underlying discriminative objectives do not necessarily require explicit modeling of how normal features vary jointly. Consequently, anomalous points may still be classified as normal if they preserve familiar semantic cues. Conversely, normal variations that alter these cues may trigger false alarms. Our feature-level diagnostics reveal both types of error, motivating anomaly evidence that captures normal feature relationships.

Reconstruction-based methods, widely used in visual anomaly detection~\cite{draem,reverse,inp}, assess feature relationships by recovering features from learned normal patterns. This provides complementary evidence for detecting missed anomalies and rejecting discriminative false alarms. However, pointwise reconstruction errors are ambiguous in sparse outdoor LiDAR scenes: normal variation can produce large residuals, while small residuals may conceal unusual feature relationships. Their magnitudes also omit temporal and spatial structure. We introduce \textbf{Structured Residual Reasoning (SRR)} to extract anomaly evidence from structured dependencies in reconstruction residuals across feature channels, time sequences, and geometric space. Normal residual statistics capture channel-wise residual distributions and cross-channel dependencies, while spatial and temporal reasoning exploits contextual consistency. To the best of our knowledge, this is the first systematic study of standalone reconstruction-based detection for LiDAR AOD. Trained without target-scene semantic labels, SRR achieves 85.13 AP, approaching the 89.00 AP of COVAL and outperforming the other evaluated segmentation-based methods.

\begin{figure*}[t]
\centering\includegraphics[width=\textwidth]{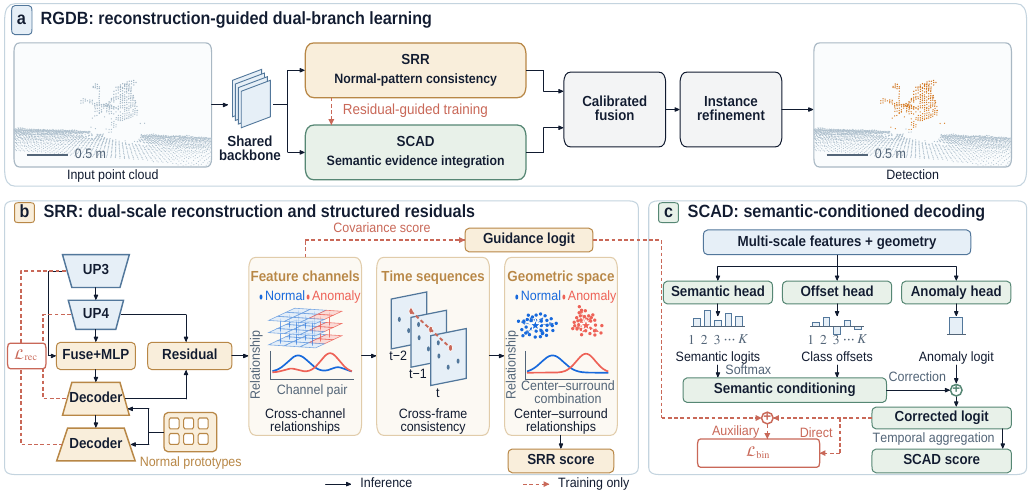}
\caption{\textbf{RGDB framework.} (a) SRR and SCAD share a backbone and provide complementary reconstruction and semantic evidence. Reconstruction guides SCAD training; calibrated fusion and instance refinement produce detections (orange) at inference. (b) SRR combines dual-scale reconstruction with structured residual reasoning across feature channels, time sequences, and geometric space. The normal residual covariance score provides guidance for SCAD training. (c) SCAD combines semantic probabilities, an anomaly logit, and class-dependent offsets to predict anomalies. Dashed arrows denote training-only paths.}
\label{fig:overview}
\end{figure*}

Reconstruction and segmentation provide complementary anomaly evidence, with each detecting anomalies missed by the other. Although their predictions can be combined through score fusion, independently trained branches cannot exploit reconstruction-derived evidence to improve semantic discrimination during training. Further analysis (Table~\ref{tab:branchcoverage}) shows that SRR recovers a broader set of objects missed by segmentation and produces fewer false alarms in normal-only scenes. These findings motivate \textbf{Reconstruction-Guided Dual-Branch (RGDB)} learning, which uses SRR's residual evidence to guide segmentation training on a shared backbone (\cref{fig:overview}). An auxiliary classification objective supervises the combined prediction, encouraging the semantic branch to complement reconstruction-based decisions, while residual-based weighting emphasizes points where reconstruction evidence is misleading or insufficient. This guided learning improves fusion performance over independently trained branches.

\begingroup
\setlength{\parskip}{0pt}
\noindent In summary, our contributions are:
\begin{itemize}[topsep=4pt,partopsep=0pt,parsep=0pt,itemsep=1pt]
\item We present the first systematic study of standalone reconstruction-based detection for LiDAR AOD and propose SRR to extract structured anomaly evidence beyond pointwise reconstruction errors.
\item We establish the complementarity between reconstruction-based and segmentation-based anomaly evidence, and propose RGDB to exploit this complementarity through reconstruction-guided segmentation learning.
\item Extensive experiments demonstrate the effectiveness of SRR without target-scene semantic labels and show that RGDB achieves state-of-the-art anomaly detection performance on the STU validation set.
\end{itemize}
\endgroup

\section{Related Work}
\paragraph{LiDAR anomalous object detection}
LiDAR anomalous object detection extends semantic segmentation to unknown objects. REAL~\cite{real} uses redundant classifiers, while APF~\cite{apf} learns class prototypes with adversarially generated unseen features. LIDO~\cite{lido} models inlier features through semantic and contrastive learning. Its segmentation head forms class prototypes, while a contrastive head models inlier feature distributions; their outputs are combined for anomaly scoring. REL~\cite{rel} scores anomalies using the energy gap between in-distribution and negative logits, and generates synthetic outliers through Point Raise perturbations. NDP~\cite{ndp} uses an attention-based distribution prior to reweight OOD scores and account for class-dependent confidence biases. Perlin noise supplies its synthetic anomaly examples. COVAL~\cite{coval} constructs paired factual and counterfactual views through visibility intervention. Reconstruction models occlusion-induced missing regions, while consistency objectives align features across the paired views.

\paragraph{Reconstruction-based Anomaly Detection}
Reconstruction methods learn to recover normal patterns. DRAEM~\cite{draem} jointly learns image reconstruction and anomaly segmentation. Reverse distillation~\cite{reverse} reconstructs pretrained features, while INP-Former~\cite{inp} guides reconstruction with image-specific normal prototypes. ReContrast~\cite{recontrast} and DNP-ConFormer~\cite{dnpconformer} adapt encoder features during reconstruction. For outdoor point clouds, Haystack~\cite{haystack} reconstructs normal scenes and feeds signed and absolute reconstruction differences into an open-set segmentation network. DOODLE~\cite{doodle} reconstructs normal semantic features with a diffusion model and uses the resulting discrepancies to amplify features for existing OOD classifiers. Both use reconstruction to support semantic anomaly prediction; our SRR instead develops structured residuals into a standalone anomaly criterion.

Beyond error magnitude, PaDiM~\cite{padim} and structured uncertainty prediction~\cite{structured} model feature and residual correlations, respectively. Masked prediction learns spatial context in images~\cite{sspcab} and point clouds~\cite{pointbert}, while road-anomaly video methods aggregate temporal evidence~\cite{roadvideo,stpaveclip}.

Building on reconstruction and contextual modeling, we study structured residuals as a standalone anomaly criterion for outdoor LiDAR. We further use normal-pattern evidence to guide the semantic branch, coupling the two criteria during training and combining their predictions at inference.

\section{Method}
\subsection{Overview}
RGDB combines Structured Residual Reasoning (SRR) and a Semantic-Conditioned Anomaly Decoder (SCAD) on a shared LIDO-pretrained sparse backbone~\cite{lido} (\cref{fig:overview}). SRR models reconstruction residuals; SCAD learns from semantic features and synthetic anomalies. SRR guides SCAD during training; inference fuses calibrated branch scores, refines instances, and feeds instance support back to point scores. Inputs are LiDAR XYZ coordinates and intensity, with poses and up to four preceding scans for temporal support. Both branches use multi-scale features and normal/synthetic training pairs; SCAD additionally uses normal-scan semantic labels.

\subsection{Synthetic Anomaly Construction}
\label{sec:synthesis}
We generate two types of synthetic anomalies from normal training scans. \textbf{Perlin Raise}, adopted from NDP~\cite{ndp}, displaces local ground points vertically according to Perlin noise, producing irregular protrusions. \textbf{Multi-shape objects} include boxes, ellipsoids, cylinders, and random convex solids placed on estimated ground. Their visible surfaces are represented by moving points along the original sensor rays to the nearer object intersections. Both constructions preserve the point count and intensity while leaving the surrounding scene unchanged.

Modification masks label altered points as synthetic anomalies and others as normal, training SCAD to distinguish the two. For SRR, normal points supervise feature reconstruction; synthetic anomalies are encouraged to retain reconstruction errors above a margin.

\subsection{Structured Residual Reasoning}
Reconstruction-based detection uses differences between original and reconstructed features as anomaly evidence. For irregular LiDAR point clouds, we reconstruct sparse voxel features and map their residuals to points for anomaly scoring. The architecture and structured residual scoring are described below.

\subsubsection{Dual-Scale Reconstruction}
We reconstruct features from the backbone's third and fourth upsampling stages, UP3 and UP4, to capture information at both scales for structured residual scoring (\cref{sec:structured-scoring}). We align UP3 features to UP4 voxel locations, lift both to $D=384$ channels, and average them into a reconstruction query:
\begin{equation}
 \ell(f)=\tfrac12[f;f;f;f],\qquad
 q_i=\tfrac12\big[\ell(f_i^3)+\ell(f_i^4)\big],
 \label{eq:query}
\end{equation}
where $f_i^3,f_i^4\in\mathbb R^{96}$ are aligned UP3 and UP4 features at query location $i$, and $\ell$ preserves the feature norm. Following INP-Former~\cite{inp}, a bottleneck MLP processes $q_i$, and eight decoder blocks reconstruct features by attending to six normal prototypes $\mathcal P=\{\mathbf p_j\}_{j=1}^6$. Shared prototypes are initialized from averaged per-frame aggregation outputs on normal scans, learned jointly with the decoder, and fixed at inference.

Let $\hat f_i^l$ denote the output of decoder block $l$. Blocks 1--4 reconstruct $\ell(f_i^4)$, while blocks 5--8 reconstruct $\ell(f_i^3)$. The auxiliary UP3 objective supervises the shared prototypes and preceding decoder blocks. We use UP4 residuals for scoring because their finer spatial resolution supports more precise point-level localization. Incorporating UP3 residuals degrades performance under the tested fusion rules (see supplementary material). We average the first four decoder outputs to compute the residual:
\begin{equation}
 \hat t_i=\tfrac14\sum\nolimits_{l=1}^{4}\hat f_i^l,\qquad
 r_i=\unit\big(\ell(f_i^4)\big)-\unit(\hat t_i),
 \label{eq:residual}
\end{equation}
where $\unit(v)=v/\|v\|_2$.

\paragraph{Reconstruction objective}
For a batch of $B$ frames, let $\mathbf f_b^3,\mathbf f_b^4$ collect backbone features at normal query locations in frame $b$, and let $\hat{\mathbf f}_b^l$ collect their reconstructions from decoder block $l$. Following INP-Former~\cite{inp}, we use a global cosine reconstruction loss:
\begin{equation}
 \mathcal L_{rec}=\frac1{8B}\sum_{b=1}^{B}\Big[\sum_{l=1}^{4}d\big(\ell(\mathbf f_b^4),\hat{\mathbf f}_b^l\big)+\sum_{l=5}^{8}d\big(\ell(\mathbf f_b^3),\hat{\mathbf f}_b^l\big)\Big],
 \label{eq:normal-reconstruction}
\end{equation}
where $d(A,B)=1-\cos(\operatorname{vec}A,\operatorname{vec}B)$ and $\operatorname{vec}$ flattens all normal-query features in a frame into one vector. The lifting operator $\ell$ acts on each query feature independently. For normal and synthetically anomalous query sets $\mathcal N$ and $\mathcal A$, let $e_i^{rec}=1-\cos(\ell(f_i^4),\hat t_i)$ denote the per-query UP4 reconstruction error. The full objective is
\begin{equation}
 \begin{aligned}
 \mathcal L_R={}&\mathcal L_{rec}
 +0.2\,\E_{i\in\mathcal N}\min_j[1-\cos(q_i,\mathbf p_j)]\\
 &+0.25\,\E_{i\in\mathcal A}[0.2-e_i^{rec}]_+.
 \end{aligned}
 \label{eq:reconstruction-training}
\end{equation}
The second term draws prototypes toward normal queries, while the third penalizes synthetic-anomaly reconstruction errors below the margin. Here $\E$ denotes a sample mean and $[x]_+=\max(x,0)$.

\paragraph{Backbone fine-tuning}
We adapt UP3 and UP4 to improve normal feature consistency and synthetic-anomaly separation while preserving pretrained features:
\begin{equation}
 \mathcal L_{feat}=\sum\nolimits_{s\in\{3,4\}}\big(\mathcal L_{fix}^{s}+\mathcal L_{compact}^{s}+\mathcal L_{sens}^{s}\big).
 \label{eq:backbone-constraint}
\end{equation}
The reference constraint $\mathcal L_{fix}^{s}$ preserves feature direction and magnitude relative to the frozen pretrained backbone. $\mathcal L_{compact}^{s}$ draws normal features toward reference directions formed from pretrained features and fixed normal-cluster centers. $\mathcal L_{sens}^{s}$ pushes synthetic features farther from normal modes than their clean counterparts. Full definitions appear in the supplementary material.

We block reconstruction gradients to the backbone to prevent feature convergence toward decoder outputs from weakening anomaly residuals.

\subsubsection{Structured Residual Scoring}
\label{sec:structured-scoring}
Pointwise residual magnitudes provide limited anomaly evidence because they overlook structure in feature, temporal, and geometric space. Cross-channel structure reveals deviations from normal feature relationships; temporal structure captures consistency across scans; geometric structure captures neighborhood support and center--surround compatibility. Structured residual scoring exploits these complementary structures for anomaly detection.

\paragraph{Normal residual covariance}
The scalar reconstruction error is proportional to the squared Euclidean norm of the 384-dimensional residual vector. This reduction preserves residual magnitude but discards its channel-wise structure. Relationships among residual channels provide additional information for distinguishing normal and anomalous points. We estimate the mean $\mu$ and covariance $\Sigma$ from normal training residuals, periodically updating them during joint training and refitting them for inference. The squared Mahalanobis score~\cite{mahalanobis1936} is:
\begin{equation}
 s_i^{cov}=(r_i-\mu)^\top(\Sigma+\epsilon I_D)^{-1}(r_i-\mu),
 \label{eq:covmain}
\end{equation}
where $I_D$ is the $D$-dimensional identity matrix, and $\epsilon=\max(10^{-8},0.1\operatorname{tr}(\Sigma)/D)$ regularizes the covariance. The score measures how strongly a residual deviates from normal channel variations and relationships. \Cref{sec:mechanism} examines how dual-scale reconstruction affects raw and covariance scoring.

\paragraph{Temporal and neighborhood support}
Sparse sampling can make anomaly scores fluctuate across scans. We use repeated observations to stabilize this evidence by aligning preceding scans with sensor poses and retaining geometrically consistent correspondences. Let $\bar s_i^H$ denote the matched historical covariance scores averaged with weights based on spatial proximity and recency. We fuse this historical support with the current score:
\begin{equation}
 s_i^T=(1-\alpha_i)s_i^{cov}+\alpha_i\bar s_i^H,
 \label{eq:temporal}
\end{equation}
where the bounded weight $\alpha_i$ increases with correspondence support. Points without valid matches retain their current scores.

Changes in sampling locations can leave points without reliable temporal correspondences. We therefore supplement correspondence-based fusion with local neighborhood support, averaging $s_i^T$ with nearby current and historical covariance scores to obtain $s_i^N$. Historical support uses unfused covariance scores to avoid recursive accumulation.

\paragraph{Center--surround compatibility}
Neighborhood aggregation stabilizes scores but does not assess feature compatibility with surroundings. We learn this relationship from normal training data by predicting the center's feature type from its neighborhood. We transform and cluster normal UP4 features into a dictionary and assign each center to its nearest type $k_i$. A context network predicts the type distribution, supervised by each center's assignment:
\begin{equation}
 \pi_i=\operatorname{softmax}\big(g_\theta(c_i)\big),\qquad
 \mathcal L_{type}=-\E_{i\in\mathcal N}\log\pi_i(k_i),
 \label{eq:context-type}
\end{equation}
where $c_i$ summarizes features and point counts in surrounding annuli, and $g_\theta$ is the context network. Training also emphasizes difficult normal samples and uses soft type distributions from similar normal neighborhoods. A low probability $\pi_i(k_i)$ indicates an unexpected center--surround relationship.

We measure this inconsistency as $-\log\pi_i(k_i)$. To assess whether it exceeds normal expectations at the current residual level, we compare it with values for normal points having similar neighborhood scores $s_i^N$, obtaining a conditional percentile $p_i\in[0,1]$. The reconstruction-branch score is
\begin{equation}
 a_i=s_i^N\exp[\beta_i(2p_i-1)].
 \label{eq:context-main}
\end{equation}
The percentile raises or lowers the score according to whether the inconsistency is above or below the conditional normal median. The bounded nonnegative gate $\beta_i$ activates only at elevated residual scores and limits the correction strength. The feature dictionary, context network, and calibration statistics are fitted on normal data and fixed at inference.

\subsection{Semantic-Conditioned Anomaly Decoder}
We use SCAD as the segmentation branch to learn discriminative anomaly evidence from normal semantics and synthetic anomalies. Following NDP~\cite{ndp}, we account for class-dependent confidence biases when constructing anomaly scores. SCAD uses LIDO-pretrained features and semantic classifier weights~\cite{lido}, combining multi-scale backbone features with point coordinates, range, and intensity.

Three heads predict semantic logits $z_i\in\mathbb R^K$, a direct anomaly logit $u_i$, and class-dependent offsets $\delta_i\in\mathbb R^K$, where $K$ is the number of known classes. With semantic probabilities $P_i=\operatorname{softmax}(z_i)$, the final anomaly logit is
\begin{equation}
 v_i=u_i+\log\sum_{k=1}^{K}P_{ik}\exp(\delta_{ik})+\rho_i.
 \label{eq:scad-main}
\end{equation}
The direct logit $u_i$ provides an initial anomaly prediction. The middle term adjusts this prediction according to the point's semantic interpretation: each $\delta_{ik}$ specifies a class-dependent correction, and $P_{ik}$ determines its contribution to the aggregation. For example, a local protrusion may resemble normal vegetation but be unusual for a road surface. If the semantic prediction favors road, a learned positive road-conditioned offset can raise its anomaly logit; if it favors vegetation, a negative vegetation-conditioned offset can lower it.

Adapting NDP's attention-based distribution prior, a learned module processes $z_i$ and $\delta_i$ to produce a bounded correction $\rho_i$. The anomaly prediction is supervised using labels from synthetic modifications, while semantic labels supervise normal-point classification.

\subsection{Dual-branch Fusion and Inference}
\paragraph{Residual-guided learning}
Our analysis in \cref{sec:branch-evidence} shows complementary anomaly detections: reconstruction recovers more objects missed by segmentation and produces fewer false positives in normal-only scenes. We therefore combine the branches and use reconstruction evidence to guide SCAD, with auxiliary classification and sample weighting encouraging compensation for reconstruction errors.

We calibrate SRR's covariance score $s_i^{cov}$ using normal-training quantiles to obtain a signed guidance logit $e_i$ compatible with SCAD's output. Positive values provide anomaly evidence, while negative values favor normality. For binary labels $y_i$, with synthetic anomalies labeled one, we supervise both SCAD's prediction and its combination with reconstruction evidence:
\begin{equation}
 \ell_i^{bin}=\tfrac12\big[\operatorname{BCE}(v_i,y_i)
 +\operatorname{BCE}(v_i+e_i,y_i)\big],
 \label{eq:guided-bce}
\end{equation}
where BCE denotes binary cross-entropy on logits. The first term directly supervises SCAD's own prediction. The second encourages compensation: positive reconstruction evidence at normal points pushes SCAD toward lower anomaly logits, while negative evidence at anomalous points pushes it toward higher logits.

We further emphasize cases where reconstruction evidence is misleading or insufficient. The covariance score is linearly normalized between its median and 99th percentile on normal training data and clipped to $[0,1]$, giving a residual level $d_i$. We assign the sample weight
\begin{equation}
 w_i=1+\lambda_w\big[(1-y_i)d_i+y_i(1-d_i)\big],
 \label{eq:guidance-weight}
\end{equation}
where $\lambda_w>0$ controls the weighting strength. High-residual normal points and low-residual anomalies receive larger weights. We compute $\mathcal L_{bin}$ by taking the weighted mean of $\ell_i^{bin}$ within each class and averaging over the classes present. Both $e_i$ and $d_i$ are detached, preventing classification gradients from passing through reconstruction evidence. Auxiliary semantic and anomaly-training losses complete the objective.

\paragraph{Calibrated fusion}
At inference, we combine the branches' final predictions. SRR provides the structured score $a_i$, while temporal aggregation of SCAD's logit $v_i$ (\cref{eq:temporal}) gives $b_i$. Normal calibration scans provide each branch's median $m_r,m_d$ and scale $s_r,s_d$, defined as the gap between its 99.5th percentile and median. The fused score is
\begin{equation}
 F_i=\tfrac12\left[\operatorname{asinh}\frac{a_i-m_r}{s_r}
 +\operatorname{asinh}\frac{b_i-m_d}{s_d}\right].
 \label{eq:fusion}
\end{equation}
Normal calibration aligns score scales; the asinh transform compresses extreme responses before equal-weight fusion.

\paragraph{Instance refinement and feedback}
Beyond pointwise anomaly detection, LiDAR AOD requires grouping anomalous points into individual objects. Common clustering methods such as DBSCAN~\cite{dbscan} can attach nearby background responses to objects or fragment objects with sparse and weak responses. We therefore introduce instance refinement to recover object extents while limiting background attachments.

To recover objects with weak or sparse responses, we group high-confidence anomaly points into spatial cores and expand them to nearby lower-score candidates. This expansion can also include surrounding background. We therefore train a membership predictor with synthetic foreground masks to distinguish object points from background, using support-plane and connectivity constraints to retain geometrically consistent members. The proportion of high-score members provides an additional criterion for rejecting unreliable candidates.

The resulting instance membership provides spatial support to refine point-level anomaly ranking. Let $I_i$ denote the instance identity of point $i$, with zero indicating background, and let $n_j=|\{i:I_i=j\}|$ denote the number of points in instance $j$. Members of instances with sufficient support receive a score increment:
\begin{equation}
 S_i=F_i+\lambda_I\,\mathbf1[I_i>0\ \land\ n_{I_i}\ge n_{min}],
 \label{eq:feedback}
\end{equation}
where $\lambda_I>0$ controls the increment and $n_{min}$ is the minimum support count. The increment is applied once after instance assignments are fixed.

\setlength{\textfloatsep}{10pt plus 2pt minus 2pt}
\setlength{\dbltextfloatsep}{10pt plus 2pt minus 2pt}
\setlength{\floatsep}{8pt plus 2pt minus 2pt}
\setlength{\intextsep}{8pt plus 2pt minus 2pt}
\captionsetup[table]{skip=4pt}
\begin{table*}[t]
\centering\small
\caption{\textbf{Official STU validation results (\%).} Published configurations are ordered by AP; gains use unrounded values. STU sem.: normal STU labels, excluding pretraining; NR: unreported; M4: Mask4Former3D; Mink: MinkowskiNet. Citations identify result sources; $^*$: REL reproduction. Published methods use single scans; SRR and RGDB additionally use up to four preceding scans with poses.}
\label{tab:comparison}
\setlength{\tabcolsep}{2.2pt}
\begin{tabular*}{\textwidth}{@{\extracolsep{\fill}}lllrrrrrrrr@{}}
\toprule
& & &\multicolumn{3}{c}{Point-Level}&\multicolumn{5}{c}{Object-Level}\\
\cmidrule(lr){4-6}\cmidrule(l){7-11}
Method & Backbone & \shortstack{STU\\sem.} & AP$\uparrow$ & FPR95$\downarrow$ & AUROC$\uparrow$ & PQ$\uparrow$ & SQ$\uparrow$ & RQ$\uparrow$ & UQ$\uparrow$ & RecallQ$\uparrow$\\\midrule
UEM$^*$~\cite{rel} & M4 & NR & 6.78 & 26.37 & 95.80 & 9.90 & 79.24 & 12.49 & 37.81 & 47.72 \\
Deep Ensemble~\cite{stu} & M4 & Yes & 6.94 & 37.34 & 90.93 & 7.27 & 79.96 & 9.10 & 14.15 & 17.70 \\
REL~\cite{rel} & M4 & Yes & 10.68 & 9.60 & 97.85 & 10.62 & 78.89 & 13.46 & 38.93 & 49.34 \\
LIDO~\cite{lido} & Mink & No & 27.53 & 34.86 & 95.05 & 21.23 & 74.82 & 28.38 & 23.88 & 31.92 \\
REAL~\cite{coval} & M4 & NR & 28.25 & 28.25 & 95.34 & 16.16 & 75.98 & 21.27 & 11.76 & 15.48 \\
NDP-EE~\cite{ndp} & M4 & Yes & 74.24 & 1.43 & 99.53 & 38.63 & 74.11 & 52.13 & 37.42 & 50.50 \\
COVAL$^{AL}$~\cite{coval} & Mink & Yes & 89.00 & 0.13 & 99.92 & 57.95 & 81.69 & 70.94 & \textbf{55.48} & \textbf{67.92} \\
\midrule
SRR (standalone) & Mink & No & 85.13 & 0.69 & 99.76 & 51.70 & 77.69 & 66.55 & 46.81 & 60.25 \\
\textbf{RGDB} & Mink & Yes & \textbf{90.03} & \textbf{0.08} & \textbf{99.94} & \textbf{60.52} & \textbf{82.86} & \textbf{73.04} & 54.45 & 65.71 \\
\bottomrule
\end{tabular*}
\end{table*}

\section{Experiments}
\subsection{Experimental Setup}
\experimentparagraph{Dataset}
We evaluate on Spotting the Unexpected (STU)~\cite{stu}, a LiDAR anomaly detection benchmark with point and instance annotations. Models are trained on its normal training split and evaluated on the 19 anomaly-validation sequences.

\experimentparagraph{Training}
We use 449 normal scans, 882 Perlin Raise examples, and 882 multi-shape examples. Joint and reconstruction-only training use identical samples. Semantic labels supervise unchanged points. RGDB is trained for 5,600 steps on two NVIDIA RTX A6000 GPUs with an effective batch size of four views. We use AdamW for the backbone and SCAD, and StableAdamW for the reconstructor, with peak learning rates of $10^{-5}$, $3\times10^{-4}$, and $10^{-4}$, respectively. All optimizers use cosine learning-rate decay and weight decay of $10^{-4}$.

\experimentparagraph{Evaluation}
We evaluate on the STU validation split following the official protocol. We report point-level AP, FPR95 and AUROC, and instance-level PQ, SQ, RQ, UQ and RecallQ, with instance matches requiring IoU $>0.5$. Continuous-scene diagnostics use STU-Cont500-v2, comprising 500 frames across 19 windows. Residual-statistics experiments use a separate 500-frame panel.

\subsection{Overall Performance and Qualitative Results}
\experimentparagraph{Benchmark comparison}
RGDB achieves 90.03 AP, 0.08 FPR95, and 60.52 PQ under the official STU validation protocol (\cref{tab:comparison}). Compared with COVAL's anomaly-logit configuration, it improves AP by 1.03 points and PQ by 2.57 points, while reducing FPR95 from 0.13\% to 0.08\%. RGDB also achieves higher SQ and RQ, indicating improved segmentation quality and instance recognition.

Without target-scene semantic labels, standalone SRR achieves 85.13 AP and 51.70 PQ, surpassing all compared methods except COVAL in both metrics. This establishes a strong reconstruction baseline. RGDB further improves AP and PQ by 4.89 and 8.82 points, respectively, and reduces FPR95 from 0.69\% to 0.08\%. Instance true positives rise from 1,387 to 1,531; false positives fall from 479 to 331.

\begin{figure*}[t]
\centering\includegraphics[width=0.92\textwidth]{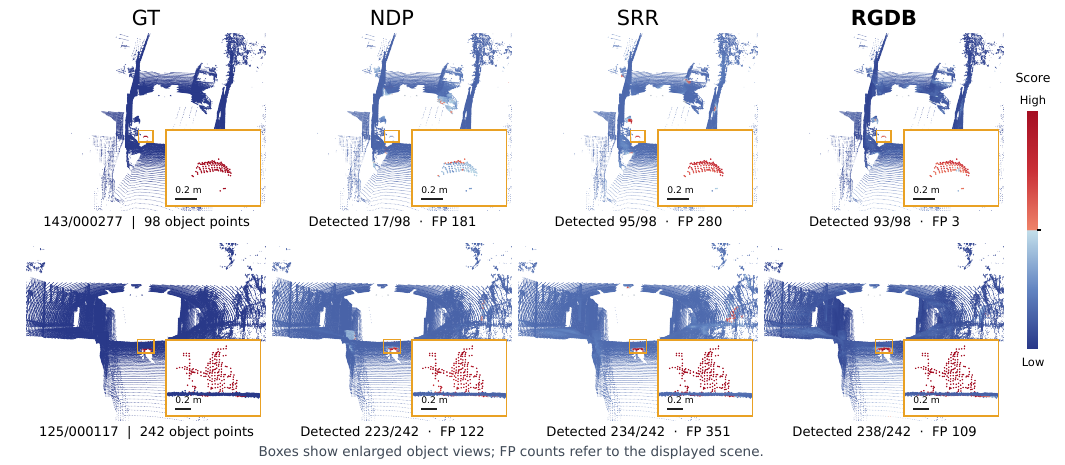}
\caption{\textbf{Qualitative comparison on two real STU scans.} GT, NDP, standalone SRR, and RGDB share measured points and views. Warm/cool colors mark scores above/below method-specific thresholds at 0.1\% background FPR across all 1,960 validation scans; color mapping is fixed across frames. Insets show boxed objects from a lower view. Counts report detected object points and background false positives in the displayed scene.}
\label{fig:qualitative}
\end{figure*}

\experimentparagraph{Qualitative comparison}
\Cref{fig:qualitative} shows more complete anomaly detection with fewer background false positives. In the first scene, RGDB detects 93 of 98 object points, compared with 17 for NDP, and produces 3 displayed background false positives, compared with 280 for standalone SRR. In the second scene, RGDB detects 238 of 242 object points with fewer background false positives than either baseline.

\subsection{Method Analysis}
\subsubsection{Complementary Anomaly Criteria}\label{sec:branch-evidence}

\begin{table}[!htb]
\centering\small
\caption{\textbf{Branch complementarity and SRR's advantages.} (a) Recovery of missed GT objects of all sizes on official STU validation. (b) False-positive points in 287 normal-only frames from the continuous 500-frame panel. Thresholds match 0.1\% background FPR over each full panel.}
\label{tab:branchcoverage}
\setlength{\tabcolsep}{5pt}
\begin{tabular}{@{}lrrr@{}}\toprule
& Objects missed & \multicolumn{2}{c}{Recovered coverage}\\
(a) Branch & by the other branch & $\geq$1 point & $\geq$50\%\\\midrule
SRR & 678 & 249 (36.73\%) & 203\\
SCAD & 497 & 68 (13.68\%) & 40\\\midrule
(b) Branch & FP points$\downarrow$ & FPR (\%)$\downarrow$ & FP reduction\\\midrule
SRR & \textbf{22,049} & \textbf{0.11137} & \textbf{24.08\%}\\
SCAD & 29,042 & 0.14669 & Reference\\\bottomrule
\end{tabular}
\end{table}

\Cref{tab:branchcoverage} compares independently trained SRR and SCAD at matched background FPR. Both recover objects missed by the other, establishing bidirectional complementarity. SRR recovers at least one point in 36.73\% of objects missed by SCAD, versus 13.68\% in reverse; coverage denotes the detected point fraction. On the continuous panel, SRR produces 24.08\% fewer false-positive points in normal-only scenes. Its broader recovery and lower normal-scene false alarms motivate reconstruction guidance. Checkpoint selection is detailed in the supplementary material.

\subsubsection{RGDB Framework Analysis and Ablation}
\experimentparagraph{Reconstruction-guided learning}
\begin{table}[!htb]
\centering\small
\caption{\textbf{Independent versus joint training.} Point-level results on official STU validation (\%). Joint denotes reconstruction-guided joint training.}
\label{tab:joint}
\setlength{\tabcolsep}{2pt}
\begin{tabular}{@{}lrrrrrr@{}}\toprule
& \multicolumn{2}{c}{SRR} & \multicolumn{2}{c}{SCAD} & \multicolumn{2}{c}{Fusion}\\
\cmidrule(lr){2-3}\cmidrule(lr){4-5}\cmidrule(l){6-7}
Training & AP$\uparrow$ & FPR95$\downarrow$ & AP$\uparrow$ & FPR95$\downarrow$ & AP$\uparrow$ & FPR95$\downarrow$\\\midrule
Independent & 83.80 & 0.3589 & 63.89 & 0.8516 & 86.79 & 0.2509\\
Joint & \textbf{84.80} & \textbf{0.3057} & \textbf{83.31} & \textbf{0.5442} & \textbf{88.47} & \textbf{0.0795}\\
\bottomrule\end{tabular}
\end{table}

Reconstruction-guided joint learning improves both branches over independent training, with a particularly large gain for the segmentation branch. As shown in \cref{tab:joint}, joint training raises SRR AP from 83.80 to 84.80 and reduces its FPR95 from 0.3589\% to 0.3057\%. SCAD gains 19.42 AP points, from 63.89 to 83.31, with FPR95 decreasing from 0.8516\% to 0.5442\%. These branch-level improvements also benefit fusion: under the same normal calibration scans and fusion rule, fusion AP increases from 86.79 to 88.47 and FPR95 decreases from 0.2509\% to 0.0795\%, a relative reduction of 68.30\%.

Intermediate-checkpoint diagnostics indicate that independent SCAD undergoes performance degradation, whereas reconstruction-guided joint training suppresses this degradation: independent SCAD drops from 87.26 to 63.89 AP, while guided SCAD improves from 81.24 to 83.31 AP over the same training interval. Training configurations and checkpoint diagnostics are detailed in the supplementary material.

\experimentparagraph{Branch fusion and instance feedback}
\begin{table}[htbp]
\centering\small
\caption{\textbf{Evidence stages on official STU validation.} Branches share the final joint checkpoint. Fusion changes point scores; feedback preserves instance assignments.}
\label{tab:stages}
\begin{tabular}{@{}lrrr@{}}\toprule
Evidence & AP$\uparrow$ & FPR95$\downarrow$ & AUROC$\uparrow$\\\midrule
Reconstruction branch & 84.80 & 0.31 & 99.77 \\
Discriminative branch & 83.31 & 0.54 & 99.82 \\
Calibrated fusion & 88.47 & 0.08 & 99.94 \\
Fusion + instance feedback & 90.03 & 0.08 & 99.94 \\
\bottomrule\end{tabular}
\end{table}
We examine the contributions and performance gains of individual components in the trained RGDB model (\cref{tab:stages}). Calibrated fusion raises AP from 84.80 for reconstruction and 83.31 for segmentation to 88.47, while reducing FPR95 from 0.31\% and 0.54\% to 0.08\%. Instance feedback further raises AP to 90.03 with unchanged instance assignments; FPR95 remains 0.08\%.

\subsubsection{SRR Analysis and Ablation}
To understand how SRR strengthens anomaly evidence beyond pointwise reconstruction errors, we analyze the contributions of its structured scoring components and dual-scale reconstruction design.

\experimentparagraph{SRR scoring components}
With the independently trained reconstructor and fitted normal statistics fixed, covariance scoring raises AP from 68.87 to 76.10 and lowers FPR95 from 9.84\% to 1.94\% (\cref{tab:srr-ablation}(a)). Temporal support adds 7.49 AP points. Local spatial components further improve detection, bringing standalone SRR to 85.13 AP and 0.69\% FPR95. Thus, residual-channel structure and temporal consistency provide the largest gains, while spatial relationships further refine anomaly evidence.

\begin{table}[t]
\centering\small
\caption{\textbf{Standalone SRR ablations (\%).} (a) Official STU validation; rows cumulatively add scoring components. Surface corrections comprise surface propagation and normal-evidence suppression. (b) UP4 residuals from two reconstruction designs on the complete 500-frame diagnostic panel before backbone adaptation. Raw denotes residual energy; Covariance uses normal residual statistics.}
\label{tab:srr-ablation}
\begin{tabular}{@{}lrr@{}}\toprule
\multicolumn{3}{@{}l@{}}{\textbf{(a) Scoring: official STU validation}}\\
Stage & AP$\uparrow$ & FPR95$\downarrow$\\\midrule
Raw residual & 68.87 & 9.84\\
Covariance & 76.10 & 1.94\\
+ Temporal support & 83.59 & 0.85\\
+ Neighborhood & 83.99 & 0.70\\
+ Center--surround & 84.50 & 0.70\\
+ Surface corrections & 85.13 & 0.69\\\midrule
\multicolumn{3}{@{}l@{}}{\textbf{(b) Reconstruction: 500-frame panel}}\\
Design & Raw AP$\uparrow$ & Covariance AP$\uparrow$\\\midrule
UP4 only & 60.35 & 63.82 \\
UP3+UP4 & 54.98 & 72.91 \\
\bottomrule\end{tabular}
\end{table}

\experimentparagraph{Dual-scale reconstruction and covariance scoring}\label{sec:mechanism}
We compare single-scale reconstruction constrained at UP4 with dual-scale reconstruction constrained at UP3 and UP4, examining their effects on raw residual scoring and covariance scoring using the same UP4 residual readout (\cref{tab:srr-ablation}(b)). Dual-scale reconstruction reduces raw-scoring AP from 60.35 to 54.98, while increasing covariance-scoring AP from 63.82 to 72.91. These results suggest that dual-scale reconstruction enriches cross-channel residual structure that covariance scoring can exploit, even when raw residual magnitude becomes less discriminative.

\section{Limitations}
RGDB may miss anomalies whose features resemble learned normal patterns, yielding small reconstruction residuals and weak discriminative responses. Normal points outside the learned distribution may cause false alarms. More diverse normal representations and complementary LiDAR geometry, image appearance, and scene semantics may help.

\section{Conclusion}
We presented RGDB for LiDAR anomalous object detection, combining semantic discrimination and normal-pattern reconstruction as complementary anomaly criteria. Semantic discrimination provides category-related evidence, while reconstruction assesses consistency with learned normal feature relationships. SRR extracts structured residual evidence across feature channels, time sequences, and geometric space, establishing a strong reconstruction baseline without semantic labels from target-scene training data. At matched background false-positive rates, SRR detects objects entirely missed by the semantic branch, while producing fewer false positives in normal-only scenes. Building on this complementarity, RGDB uses reconstruction-derived normality cues to guide SCAD during joint learning, improving both branches and suppressing the performance degradation observed in independently trained SCAD. Calibrated fusion and instance refinement combine their predictions at inference. Experiments demonstrate that RGDB achieves state-of-the-art performance in point-level anomaly detection and instance segmentation on the STU validation set.

\clearpage
{\small\bibliographystyle{ieeenat_fullname}\bibliography{references}}

\begin{thebibliography}{22}
\providecommand{\natexlab}[1]{#1}
\providecommand{\url}[1]{\texttt{#1}}
\expandafter\ifx\csname urlstyle\endcsname\relax
  \providecommand{\doi}[1]{doi: #1}\else
  \providecommand{\doi}{doi: \begingroup \urlstyle{rm}\Url}\fi

\bibitem[Cen et~al.(2022)Cen, Yun, Zhang, Cai, Luan, Wang, Liu, and Tang]{real}
Jun Cen, Peng Yun, Shiwei Zhang, Junhao Cai, Di Luan, Michael~Yu Wang, Ming
  Liu, and Mingqian Tang.
\newblock Open-world semantic segmentation for {LiDAR} point clouds.
\newblock In \emph{ECCV}, 2022.

\bibitem[Defard et~al.(2021)Defard, Setkov, Loesch, and Audigier]{padim}
Thomas Defard, Aleksandr Setkov, Angelique Loesch, and Romaric Audigier.
\newblock {PaDiM}: A patch distribution modeling framework for anomaly
  detection and localization.
\newblock In \emph{Pattern Recognition. ICPR International Workshops and
  Challenges}, pages 475--489, 2021.

\bibitem[Deng and Li(2022)]{reverse}
Hanqiu Deng and Xingyu Li.
\newblock Anomaly detection via reverse distillation from one-class embedding.
\newblock In \emph{CVPR}, pages 9737--9746, 2022.

\bibitem[Ester et~al.(1996)Ester, Kriegel, Sander, and Xu]{dbscan}
Martin Ester, Hans-Peter Kriegel, J{\"u}rg Sander, and Xiaowei Xu.
\newblock A density-based algorithm for discovering clusters in large spatial
  databases with noise.
\newblock In \emph{Proceedings of the Second International Conference on
  Knowledge Discovery and Data Mining}, pages 226--231, 1996.

\bibitem[Faulkner et~al.(2025)Faulkner, Haub, Ratcliffe, and Chin]{haystack}
Ryan Faulkner, Luke Haub, Simon Ratcliffe, and Tat-Jun Chin.
\newblock Finding outliers in a haystack: Anomaly detection for large
  pointcloud scenes.
\newblock In \emph{ACPR}, 2025.

\bibitem[Guo et~al.(2023)Guo, Lu, Jia, Zhang, and Li]{recontrast}
Jia Guo, Shuai Lu, Lize Jia, Weihang Zhang, and Huiqi Li.
\newblock {ReContrast}: Domain-specific anomaly detection via contrastive
  reconstruction.
\newblock In \emph{Advances in Neural Information Processing Systems}, 2023.

\bibitem[He et~al.(2026)He, Huang, Wang, Yang, and Li]{stpaveclip}
Siyuan He, Yuchun Huang, Chen Wang, Feng Yang, and Yifan Li.
\newblock {ST-PaveCLIP}: A spatio-temporal vision--language framework for road
  anomaly segmentation in images and videos.
\newblock \emph{Remote Sensing}, 18\penalty0 (17):\penalty0 2922, 2026.

\bibitem[Li and Dong(2023)]{apf}
Jianan Li and Qiulei Dong.
\newblock Open-set semantic segmentation for point clouds via adversarial
  prototype framework.
\newblock In \emph{CVPR}, pages 9425--9434, 2023.

\bibitem[Li et~al.(2026{\natexlab{a}})Li, Liu, Lin, Shen, Wang, and
  Fu]{dnpconformer}
Luhu Li, Bin Liu, Bowen Lin, Zihan Shen, Chengwei Wang, and Shujun Fu.
\newblock Diverse normal prototypes-guided contrastive reconstruction for
  medical anomaly detection.
\newblock In \emph{Medical Image Computing and Computer Assisted Intervention
  -- MICCAI 2026}, 2026{\natexlab{a}}.

\bibitem[Li et~al.(2025)Li, Xiang, Ao, West, and Khoshelham]{rel}
Zizhao Li, Zhengkang Xiang, Jiayang Ao, Joseph West, and Kourosh Khoshelham.
\newblock Relative energy learning for {LiDAR} out-of-distribution detection.
\newblock \emph{arXiv preprint arXiv:2511.06720}, 2025.

\bibitem[Li et~al.(2026{\natexlab{b}})Li, Xiang, Ao, Liu, West, and
  Khoshelham]{ndp}
Zizhao Li, Zhengkang Xiang, Jiayang Ao, Feng Liu, Joseph West, and Kourosh
  Khoshelham.
\newblock Neural distribution prior for {LiDAR} out-of-distribution detection.
\newblock In \emph{CVPR}, pages 3035--3045, 2026{\natexlab{b}}.

\bibitem[Luo et~al.(2025)Luo, Cao, Yao, Zhang, Lou, Cheng, Shen, and Yu]{inp}
Wei Luo, Yunkang Cao, Haiming Yao, Xiaotian Zhang, Jianan Lou, Yuqi Cheng,
  Weiming Shen, and Wenyong Yu.
\newblock Exploring intrinsic normal prototypes within a single image for
  universal anomaly detection.
\newblock In \emph{CVPR}, pages 9974--9983, 2025.

\bibitem[Mahalanobis(1936)]{mahalanobis1936}
Prasanta~Chandra Mahalanobis.
\newblock On the generalised distance in statistics.
\newblock \emph{Proceedings of the National Institute of Sciences of India},
  2\penalty0 (1):\penalty0 49--55, 1936.

\bibitem[Mosco et~al.(2026)Mosco, Fusaro, and Pretto]{lido}
Simone Mosco, Daniel Fusaro, and Alberto Pretto.
\newblock Learning to identify out-of-distribution objects for {3D LiDAR}
  anomaly segmentation.
\newblock In \emph{CVPR}, pages 17204--17214, 2026.

\bibitem[Nekrasov et~al.(2025)Nekrasov, Burdorf, Worrall, Leibe, and
  Perez]{stu}
Alexey Nekrasov, Malcolm Burdorf, Stewart Worrall, Bastian Leibe, and Julie
  Stephany~Berrio Perez.
\newblock Spotting the unexpected ({STU}): A {3D LiDAR} dataset for anomaly
  segmentation in autonomous driving.
\newblock In \emph{CVPR}, pages 11875--11885, 2025.

\bibitem[Oh et~al.(2026)Oh, Kim, We, Jeong, Chae, and Yoon]{doodle}
Changgyoon Oh, Hyeonseong Kim, Daehyun We, Jongoh Jeong, Yujeong Chae, and
  Kuk-Jin Yoon.
\newblock {DOODLE}: Diffusion-based out-of-distribution learning for open-set
  {LiDAR} semantic segmentation.
\newblock In \emph{Proceedings of the IEEE/CVF Winter Conference on
  Applications of Computer Vision (WACV)}, pages 2273--2283, 2026.

\bibitem[Rai et~al.(2026)Rai, Karthik, Georgescu, Caputo, Masone, and
  Akata]{roadvideo}
Shyam~Nandan Rai, Shyamgopal Karthik, Mariana-Iuliana Georgescu, Barbara
  Caputo, Carlo Masone, and Zeynep Akata.
\newblock Road obstacle video segmentation.
\newblock In \emph{Pattern Recognition. DAGM GCPR 2025}, pages 186--201, 2026.

\bibitem[Ristea et~al.(2022)Ristea, Madan, Ionescu, Nasrollahi, Khan, Moeslund,
  and Shah]{sspcab}
Nicolae-C\u{a}t\u{a}lin Ristea, Neelu Madan, Radu~Tudor Ionescu, Kamal
  Nasrollahi, Fahad~Shahbaz Khan, Thomas~B. Moeslund, and Mubarak Shah.
\newblock Self-supervised predictive convolutional attentive block for anomaly
  detection.
\newblock In \emph{CVPR}, pages 13576--13586, 2022.

\bibitem[Vodenicharski et~al.(2022)Vodenicharski, McDermott, Webber, Introini,
  Cicuta, Bowman, Simpson, and Campbell]{structured}
Boyko Vodenicharski, Samuel McDermott, Katherine~M. Webber, Viola Introini,
  Pietro Cicuta, Richard Bowman, Ivor J.~A. Simpson, and Neill D.~F. Campbell.
\newblock Cell anomaly localisation using structured uncertainty prediction
  networks.
\newblock In \emph{Medical Imaging with Deep Learning}, pages 1285--1300, 2022.

\bibitem[Yang et~al.(2026)Yang, Liu, Tan, Shen, Shen, Zhu, and Hu]{coval}
Longyu Yang, Jun Liu, Yap-Peng Tan, Fumin Shen, Heng~Tao Shen, Xiaofeng Zhu,
  and Ping Hu.
\newblock Counterfactual occlusion-aware learning via visibility intervention
  for {LiDAR} anomaly detection.
\newblock In \emph{ICML}, pages 145532--145548, 2026.

\bibitem[Yu et~al.(2022)Yu, Tang, Rao, Huang, Zhou, and Lu]{pointbert}
Xumin Yu, Lulu Tang, Yongming Rao, Tiejun Huang, Jie Zhou, and Jiwen Lu.
\newblock {Point-BERT}: Pre-training {3D} point cloud transformers with masked
  point modeling.
\newblock In \emph{CVPR}, pages 19313--19322, 2022.

\bibitem[Zavrtanik et~al.(2021)Zavrtanik, Kristan, and Sko\v{c}aj]{draem}
Vitjan Zavrtanik, Matej Kristan, and Danijel Sko\v{c}aj.
\newblock {DRAEM}: A discriminatively trained reconstruction embedding for
  surface anomaly detection.
\newblock In \emph{ICCV}, pages 8330--8339, 2021.

\end{thebibliography}
\end{document}